\documentclass[letterpaper,10pt,conference]{ieeeconf}
\IEEEoverridecommandlockouts
\usepackage{amsmath,amssymb}
\usepackage{graphicx}
\usepackage{microtype}
\usepackage{booktabs}
\usepackage{url}
\usepackage{xcolor}
\usepackage{colortbl}
\usepackage{tabularx}
\usepackage[hidelinks]{hyperref}

\newcommand{\vect}[1]{\boldsymbol{#1}}
\newcommand{\mat}[1]{\mathbf{#1}}

\definecolor{amandapurple}{RGB}{128,0,192}

\definecolor{amandablue}{RGB}{0,60,125}

\definecolor{revisionteal}{RGB}{0,105,105}

\newcommand{\bestresult}[1]{\textbf{#1}}

\title{\LARGE \bf
Weakly Supervised Discovery of Motion Styles \\ for Compositional Robot Control}
\author{Heedo Woo \and[\hspace{1.5em}] Antonio Marino \and[\hspace{1.5em}] Eduardo Sebasti\'{a}n \and[\hspace{1.5em}] Amanda Prorok%
\thanks{All authors are with the Department of Computer Science and Technology, University of Cambridge. Email: \texttt{\{hw527,\protect\allowbreak am3507,\protect\allowbreak es2121,\protect\allowbreak asp45\}@cam.ac.uk}.}%
\thanks{This work is supported by a Leverhulme Trust Research Project Grant. We gratefully acknowledge their support.}}

\begin{document}
\maketitle
\thispagestyle{empty}
\pagestyle{empty}

\definecolor{mygreen}{rgb}{0.1,0.86,0.16}\def\mygreen#1{{\textcolor{mygreen}{#1}}}

\newcommand{\AM}[1]{\mygreen{\footnote{\mygreen{[AM]: #1}}}}
\newcommand{\AMl}[1]{\mygreen{[AM]: #1}}

\newcommand{\reviewsectionbreak}{\clearpage}

\begin{abstract}
When a robot possesses more degrees of freedom than a task requires, multiple trajectories can achieve the same objective using distinct motion styles. Compositional robot control aims to preserve a selected style while simultaneously executing an independent task command, effectively balancing these potentially competing objectives.
In this paper, we present CODES, a framework for compositional robot control that discovers a coherent style repertoire from diverse recordings and composes it with independent commands. First, motion retargeting maps the recordings to motions executable by the target robot; next, weakly supervised consensus encoding and density-based clustering identify styles within and across recordings without a prescribed count; finally, reinforcement learning trains a policy to compose the discovered styles with independent commands.
We evaluate CODES on a Unitree Go2 quadruped across two simulators and physical hardware. Compared to baselines, CODES significantly improves kinematic coherence. Furthermore, our controller successfully tracks velocity commands that generalize beyond the support of the recorded data, exhibiting robust and distinguishable style characteristics during both cross-simulator and real-world hardware deployments.

 Project website: \href{https://www.proroklab.org/compositional-motion-styles/}{proroklab.org/compositional-motion-styles}
\end{abstract}

\begin{keywords}
Compositional robot control, motion style discovery, weakly supervised learning, reinforcement learning.
\end{keywords}

\section{Introduction}
\label{sec:introduction}

Robot motion control is commonly formulated in terms of task objectives, such as target velocities, poses, or trajectories. However, a given objective typically admits many executions that differ in motion characteristics, including posture, contact patterns, and coordination. These characteristics, referred to here as \emph{motion styles}, influence aspects such as clearance, interaction with the environment, and predictability~\cite{margolis2023walk,allgaits2025}. This freedom is particularly relevant when a robotic system has more degrees of freedom than those that the task constrains. The task command alone does not select these characteristics. A \emph{compositional} robot controller therefore accepts a style reference and a task command as independently selectable inputs, so that the requested execution can change without changing the task objective.

\begin{figure*}[!t]
 \centering
 \IfFileExists{figures/pipeline.pdf}{%
  \includegraphics[width=\textwidth]{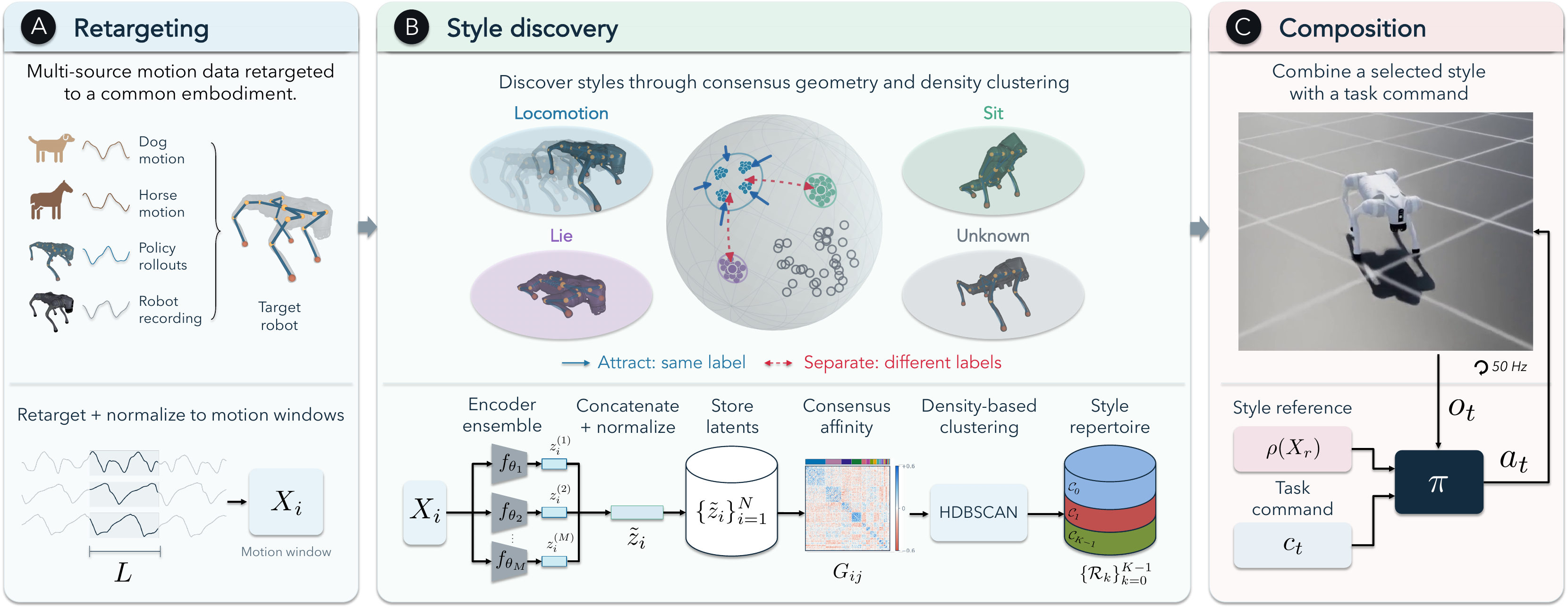}%
 }{%
  \fbox{\begin{minipage}[c][40mm][c]{0.94\textwidth}
   \centering
   Style-discovery pipeline figure.\\[1ex]
   The image is omitted from the LaTeX-only checkout.\\
   Add \texttt{figures/pipeline.pdf} to display it.
  \end{minipage}}%
 }
 \caption{\textbf{Learning and composing motion styles for quadruped control.} (A) Motions are retargeted to a common representation compatible with the robot's kinodynamics; (B) a weakly-supervised method based on consensus affinity and HDBSCAN identifies a style repertoire; and (C) a policy trained with reinforcement learning composes a style reference from the repertoire and an independent task command for robot control. This makes recorded motion characteristics available under independently selected commands.}
 \label{fig:style_discovery_pipeline}
\end{figure*}

To compose motion styles and task commands, some representation of the styles is needed. Motion style characteristics can be specified explicitly through coupled oscillators, such as in AllGaits~\cite{allgaits2025}, or parameterized via posture commands, such as in Walk These Ways~\cite{margolis2023walk}, allowing selection within predefined families of styles. As an alternative to manual parameterizations, motion recordings offer more flexibility for learning motion structure~\cite{mann2018,vqstyle2026}, but entangle embodiments, regimes, and constraints. In computational ethology, B-SOiD~\cite{hsu2021bsoid}, VAME~\cite{luxem2022vame}, and Keypoint-MoSeq~\cite{weinreb2024keypointmoseq} identify recurring patterns from pose sequences without predefined taxonomies. A single recording may contain several styles, and similar styles may recur across recordings. Segmentation alone does not identify the characteristics that should persist as task commands change.

On the control side, reference-based reinforcement learning uses demonstrations through explicit tracking (DeepMimic~\cite{deepmimic2018}) or learned adversarial motion rewards (AMP~\cite{peng2021amp}). Multi-behavior controllers select among motions using separate adversarial motion priors (Multi-AMP~\cite{vollenweider2023multiamp}), continuous embeddings~\cite{peng2022ase,tessler2023calm}, or discrete codes~\cite{zhu2023ncp,han2024lifelike}. Coarse annotations can nevertheless group distinct execution patterns under one label; diverse execution therefore does not ensure a distinguishable, \emph{coherent} style representation.

Consequently, neither raw recording boundaries nor existing annotations alone define an adequate repertoire. For successful compositional control, styles extracted from data must be \emph{coherent} and independently selectable from task commands. This separation parallels Walk Like Dogs~\cite{kang2025steerable}, which pairs locomotion modes independently with steering commands. Kinematic coherence requires windows in the same style group to share motion characteristics~(e.g., posture, joint-velocity patterns, and periodicity), with windows from different groups remaining distinguishable. The controller must exhibit both \emph{command} and \emph{style fidelity}, following the task command while preserving the selected style's motion characteristics.

These considerations motivate CODES (Composition Of Discovered Expressive Styles), which identifies motion styles from recordings and trains a policy to execute them under independently specified task commands (Sec.~\ref{sec:problem}). Recordings are first expressed on the target robot, providing executable motion examples from which candidate styles are identified by a weakly supervised discovery method; these styles are then exposed to the policy as a separate control input (Sec.~\ref{sec:method}). We evaluate the framework in a quadruped case study on the Unitree Go2 (Sec.~\ref{sec:case_study}). Ablations and comparisons with alternative decomposition methods show a compact repertoire with higher kinematic coherence than the no-coarse-guidance and VQ-VAE baselines (Sec.~\ref{sec:controller_evaluation}). We further assess the robot motion controller in two simulators and real hardware deployments, showing that the resulting policy achieves command and style fidelity across a broad range of motions.

\section{Problem Formulation}
\label{sec:problem}

Consider a target robot whose state is $\vect x\in\mathbb R^{d_x}$ and a set of state trajectories recorded from the same robot, other robots, or other embodiments, such as animals. The recorded trajectories are organized into windows
\mbox{$\mat S_i=[\vect s_{i,1}^{\top};\ldots;\vect s_{i,L}^{\top}] \in\mathbb R^{L\times d_s}$}
of $L$ consecutive recorded states $\vect s\in\mathbb R^{d_s}$, with $;$ denoting vertical concatenation. The target and recorded state representations might not coincide and can have different dimensions, \mbox{$d_x\ne d_s$}.
The recorded data might also carry a coarse label $y_i\in\mathcal Y$, optionally accompanied by an annotation confidence $\xi_i\in[0,1]$; hard labels use $\xi_i=1$. The vocabulary $\mathcal Y$ contains broad motion categories, and \emph{unknown} denotes missing coarse-label supervision rather than an additional motion category and uses $\xi_i=0$. The motion collection is
\begin{equation}
 \mathcal D=\{(\mat S_i,y_i)\}_{i=1}^{N},
 \label{eq:motion_collection}
\end{equation}
where $N$ is the number of windows. The coarse labels only specify priors available from human expertise and provide weak supervision: they identify broad motion categories rather than the fine-grained styles or window assignments to be inferred.
Within the recorded trajectory windows, we can identify a set of motion characteristics shared by multiple windows. In this paper, we denote such characteristics as a \emph{motion style}.

Styles are therefore represented by a partial grouping of the collection, which assigns each window a label \mbox{$k_i\in\{-1,0,\ldots, K-1\}$}, where $k_i=-1$ denotes a window without sufficient support for any style. For each style $k$, we define the group $\mathcal C_k$ and a nonempty reference pool $\mathcal R_k$ of observed windows from that group:
\begin{equation}
 \mathcal C_k=\{i:k_i=k\},
 \qquad
 \mathcal R_k\subseteq\{\mat \rho(\mat S_i):i\in\mathcal C_k\}.
 \label{eq:reference_pool}
\end{equation}
Each element $\vect r\in\mathcal R_k$ is an observed example of style $k$, with $\rho(\mat S_i)$ a representation of $\mat S_i$, the union \mbox{$\mathcal R=\bigcup_{k=0}^{K-1}\mathcal R_k$} is the reference bank, and the groups $\{\mathcal C_k\}_{k=0}^{K-1}$ form the \emph{style repertoire}. The repertoire is \textbf{coherent} if windows within a group share motion characteristics, windows from different groups are distinguishable, and consecutive windows of a recording share an assignment while the underlying motion pattern persists. A coherent grouping of $\mathcal D$ need not be unique, and the metrics to measure it are task- and robot-dependent. Neither the styles nor their number $K$ is known a priori.

Once a coherent style repertoire is established, it can be used for motion control. At time step $t$, the robot receives an observation $\vect o_t$ and a task command $\vect u_t\in\mathcal U$, where $\mathcal U$ is the set of admissible commands. A style is selected by providing a reference $\vect r_t\in\mathcal R_k$, and the style-conditioned policy $\pi$ produces the action
\begin{equation}
 \vect a_t\sim\pi\bigl(\,\cdot\mid\vect o_t,\vect u_t,\vect r_t\bigr).
 \label{eq:policy}
\end{equation}
The command specifies what the execution must achieve, and the reference specifies how the motion is performed.

Let $\Bar{\mat X}_{\vect r,\vect u}\in\mathbb R^{L\times d_x}$ denote a window of states executed by the robot for some $(\vect r,\vect u)$. The achieved command sequence is
$\Bar{\vect u}=\operatorname{cmd}(\Bar{\mat X}_{\vect r,\vect u})$, where \mbox{$\operatorname{cmd}:\mathbb R^{L\times d_x}\to\mathcal U$} maps an executed motion window to its sequence of command-relevant quantities. Deviations are measured by
a command distance $d_{\mathrm{cmd}}$ and a style distance \mbox{$d_{\mathrm{style}}$}, with tolerances $\varepsilon_{\mathrm{cmd}}>0$ and $\varepsilon_{\mathrm{style}}>0$. Ideally, $d_{\mathrm{style}}$ compares the style characteristics and disregards differences attributable to the command.

The policy
displays \textbf{command fidelity} for a reference $\vect r$ if it tracks every admissible command,
\begin{equation}
 d_{\mathrm{cmd}}\bigl(\Bar{\vect u},\vect u\bigr)
 \le\varepsilon_{\mathrm{cmd}}
 \quad\forall\,\vect u\in\mathcal U,
 \label{eq:commandability}
\end{equation}
and \textbf{style fidelity} for a command $\vect u$ if the executed motion preserves every selectable style,
\begin{equation}
 d_{\mathrm{style}}\bigl(\Bar{\mat X}_{\vect r,\vect u},\vect r\bigr)
 \le\varepsilon_{\mathrm{style}}
 \quad\forall\,\vect r\in\mathcal R.
 \label{eq:expressibility}
\end{equation}
Since the commands under which style $k$ was recorded typically cover only a subset of $\mathcal U$, the two properties jointly require generalization beyond the recorded style-command combinations.

Given a motion collection $\mathcal D$, our \textbf{goal} is to find a coherent style repertoire $\{\mathcal C_k\}_{k=0}^{K-1}$ and a policy $\pi$ that displays command and style fidelity for all $\vect r\in\mathcal R$ and $\vect u\in\mathcal U$.

\section{Proposed Framework}
\label{sec:method}

In this section, we detail CODES, our proposed framework for integrating motion representation, style discovery, and compositional control. As outlined in Figure~\ref{fig:style_discovery_pipeline}, the pipeline begins by mapping motion recordings onto the target robot. Weakly-supervised encoders are then used to construct a consensus representation, from which the style repertoire and its reference bank are identified via density-based clustering. Finally, a style-conditioned policy combines the selected style with an independently specified task command. We detail each step in the following subsections.

\subsection{Retargeting to the Target Robot}

To obtain comparable, executable motion examples, CODES maps each recorded window through a preparation map $\mathcal T_i$,
\begin{equation}
 \mat X_i=\mathcal T_i(\mat S_i)\in\mathbb R^{L\times d_x}.
 \label{eq:target_window}
\end{equation}
The preparation map has two requirements: all outputs use common state coordinates, units, and sampling rate, and each output is kinodynamically feasible for the target robot. These requirements make different sources comparable and ground the discovered styles in executable motion, supporting commandability in Eq.~\eqref{eq:commandability}. Recordings that already satisfy them use the identity map. CODES otherwise does not prescribe how $\mathcal T_i$ is realized; depending on the source and target embodiments, it may use inverse-kinematic~\cite{peng2020animals} or kino-dynamic~\cite{kang2025steerable} mappings, key-body correspondence~\cite{araujo2025gmr}, interaction meshes~\cite{yang2025omniretarget}, or physics-based imitation~\cite{deepmimic2018,mimickit2025}.

\subsection{Weakly-Supervised Motion Representation}
\label{sec:discovery}
\label{sec:discovery_requirements}

A style representation should describe motion over time while preserving differences within broad behavior categories. Fixed-duration windows capture local coordination without treating an entire recording as one style. We use a compact one-dimensional temporal convolutional network (TCN)~\cite{bai2018tcn} to map each window to a fixed-dimensional code that can be compared for clustering and supplied to control.
Supervised contrastive learning can admit solutions in which each class collapses to a single representation~\cite{graf2021supcon}, discarding information needed for finer distinctions~\cite{fu2022classcollapse}. Reconstruction and instance contrast retain motion detail, while balanced coarse-label guidance organizes the representation without prescribing the final groups. The labels enter the training objective rather than the encoder input.

Specifically, we train an ensemble of $M$ encoders using the same data and objective but different random initializations. Each encoder applies three TCN blocks to the motion window, followed by temporal mean and maximum pooling and an MLP, with parameters $\theta_m$. Its $d$-dimensional output is normalized to the unit sphere
\begin{equation}
  \vect z_i^{(m)}=f_{\theta_m}(\mat X_i)\in\mathbb R^d,
  \qquad \|\vect z_i^{(m)}\|_2=1,
  \label{eq:encoder}
\end{equation}
Here, $m\in\{1,\ldots,M\}$ indexes the encoders. During training, each encoder and its decoder use a loss that combines reconstruction and contrastive terms,
\begin{equation}
 \mathcal L
 =\mathcal L_{\mathrm{recon}}
 +\lambda_{\mathrm{inst}}\mathcal L_{\mathrm{robust}}
 +\lambda_+\mathcal L_{\mathcal Y}^{+}+\lambda_-\mathcal L_{\mathcal Y}^{-},
 \label{eq:representation_loss}
\end{equation}
with $\lambda_{\bullet}$ nonnegative weights. Reconstruction preserves the motion within each window; instance contrast encourages robustness to perturbations while distinguishing windows; and coarse-label attraction and separation provide broad behavioral guidance.

The decoder reconstructs the clean, normalized window $\mat X$ from each of its two augmented views, producing $\widehat{\mat X}$. The weighted reconstruction loss is
\begin{equation}
  \mathcal L_{\mathrm{recon}}
  = \lambda_x\bigl\langle\ell_{\mathrm{sL1}}(\widehat{\mat X},\mat X)\bigr\rangle + \lambda_\Delta\bigl\langle\ell_{\mathrm{sL1}}(\Delta\widehat{\mat X},\Delta\mat X)\bigr\rangle.
 \label{eq:reconstruction}
\end{equation}
Here $\ell_{\mathrm{sL1}}$ is the mean element-wise smooth-$L_1$ error, $\Delta$ denotes the difference between consecutive frames, and brackets average over sampled windows and both augmented views. The terms preserve configurations and frame-to-frame changes, as in motion reconstruction~\cite{t2mgpt2023} and velocity reconstruction~\cite{vqstyle2026}.

$\mathcal{L}_{\mathrm{robust}}$ uses symmetric NT-Xent~\cite{simclr2020} with temperature-scaled cosine similarity between unit-normalized codes. Two views of each window, independently perturbed by channel gain, channel dropout, and Gaussian noise, form the positive pair; views of other windows are negatives, promoting robustness but potentially separating similar motions.

We address this drawback with the last two loss terms using the labels. Let indices $a$ and $b$ denote augmented views with inherited coarse labels $y$ and annotation confidences $\xi \in [0,1]$. The attraction term uses supervised contrastive learning~\cite{khosla2020supcon}, with views of other windows sharing the same known label as positives. Positive contributions are weighted by \(\min(\xi_a,\xi_b)\) to obtain \(\ell_a^{\mathrm{sup}}\), while all other views in the batch remain in the contrastive denominator. This gives the attraction loss
\begin{equation}
 \mathcal L_{\mathcal Y}^{+}
 =\frac{\sum_a\xi_a\alpha_{y_a}\ell_a^{\mathrm{sup}}}{\sum_a\xi_a},
 \label{eq:coarse_attraction}
\end{equation}
where the multiplier $\alpha_y$ scales inversely with class support to prevent over-compressing frequent, coarse classes while boosting rare ones. 

Conversely, the separation term penalizes different-label pairs only if their cosine similarity exceeds a margin $\mu$:
\begin{equation}
 \mathcal L_{\mathcal Y}^{-}
 =\frac{\sum_{a,b}w_{ab}[\vect z_a^\top\vect z_b-\mu]_+^2}{\sum_{a,b}w_{ab}},
 \label{eq:coarse_separation}
\end{equation}
where $[x]_+=\max(x,0)$ and pair weights \mbox{$w_{ab}=\xi_a\xi_b/\sqrt{n_{y_a}n_{y_b}}$} depend on class window counts $n_y$. This margin ignores moderately similar pairs, shaping the representation without overly constraining downstream clustering. Unknown windows contribute to reconstruction and instance contrast, but not to the coarse-label attraction or separation terms.

\subsection{Consensus Geometry and Style Decomposition}

Because density clustering is sensitive to local neighborhoods, which vary with random initialization, we aggregate affinities across the encoder ensemble to form a robust geometry~\cite{strehl2002ensemble}. For windows $i$ and $j$, we concatenate the $M$ encoder outputs into a unit-norm consensus code, ensuring its inner product exactly equals the mean affinity:
\begin{equation}
  \begin{aligned}
  \widetilde{\vect z}_i=\frac{1}{\sqrt M}
  [\vect z_i^{(1)};\ldots;\vect z_i^{(M)}]~\in\mathbb R^{Md},\\
  \quad  G_{ij}=\widetilde{\vect z}_i^\top\widetilde{\vect z}_j=\frac{1}{M}\sum_{m=1}^{M}(\vect z_i^{(m)})^\top\vect z_j^{(m)}.
 \end{aligned}
 \label{eq:consensus}
\end{equation}
Because neither the number of styles $K$ nor the assignment of every window is known in advance, we apply \mbox{HDBSCAN}~\cite{hdbscan2017}, a hierarchical density-based clustering method, to the consensus geometry. Since consensus codes have unit norm, the Euclidean distance used for density clustering is entirely determined by the mean cosine affinity $\|\widetilde{\vect z}_i-\widetilde{\vect z}_j\|_2=\sqrt{2-2G_{ij}}$.

HDBSCAN operates exclusively on these consensus codes without coarse labels, recording identities, or manual merging, allowing the number of styles $K$ to emerge naturally from the data. This produces the partial assignment \mbox{$k_i\in\{-1,0,\ldots,K-1\}$} of Section~\ref{sec:problem}, where $k_i=-1$ marks windows lacking sufficient density support. Each discovered group $\mathcal C_k$ defines a reference pool $\mathcal R_k$ from its consensus codes, collectively establishing the reference bank $\mathcal R$ used for downstream control.

\subsection{Style-Conditioned Composition}

CODES conditions the policy in Eq.~\eqref{eq:policy} on a selected consensus code, $\vect r_t=\rho(\mat S_r)=\widetilde{\vect z}_r\in\mathcal R_k$, and an independent task command $\vect u_t$. Codes are computed from retargeted windows $\mat X_r=\mathcal T_r(\mat S_r)$ through Eqs.~\eqref{eq:encoder} and~\eqref{eq:consensus}. The encoders are frozen, and reference codes are stored before policy training.

Consensus similarity compares motion windows; a conditional adversarial motion prior (AMP)~\cite{peng2021amp} supplies a training reward for individual transitions. Its discriminator distinguishes reference from policy transitions, conditioned on the selected code. Reference transitions are sampled independently of the execution phase, so no frame-by-frame alignment or online window encoding is required.

Command tracking and reference compatibility can favor different executions. We train the policy with proximal policy optimization (PPO) using the gated reward
\begin{equation}
 r_t=r_t^{\mathrm{aux}}+w_{\mathrm{AMP}}r_t^{\mathrm{AMP}}
     +w_{\mathrm{cmd}}\,g(r_t^{\mathrm{AMP}})\,r_t^{\mathrm{cmd}},
 \label{eq:composed_reward}
\end{equation}
where $r_t^{\mathrm{AMP}}\in[0,1]$ rewards transitions the discriminator judges consistent with the selected reference data, $r_t^{\mathrm{cmd}}$ rewards command tracking, and $r_t^{\mathrm{aux}}$ contains robot-specific regularization. The logistic gate is
\begin{equation}
 g(r)=\nu+\frac{1-\nu}{1+\exp\!\left(-(r-\delta_g)/T_g\right)},
 \label{eq:style_gate}
\end{equation}
with floor $0<\nu<1$, midpoint $\delta_g$, and width $T_g>0$. Higher AMP rewards smoothly increase the tracking contribution, while the floor retains a tracking incentive at low scores. The additive AMP term rewards reference compatibility independently of tracking accuracy.

During training, references are paired with recording-derived commands and commands sampled independently from $\mathcal U$, exposing the policy to combinations absent from the recordings. References and commands remain independently selectable at execution.
\section{Case Study: Quadruped Motion Control}
\label{sec:case_study}

We instantiate and evaluate our framework using a quadrupedal robot as the target platform. Quadrupeds possess a highly redundant and expressive action space, making them an ideal testbed for style discovery and compositional robot control.

The collection combines policy rollouts based on Kine2Go reference motions~\cite{kine2go2026} originating from dogs, horses, and other robot platforms, with motion-capture recordings of a Unitree Go2 teleoperated using its built-in controller. The motion styles are learned over the robot state projected into its local frame with $d_x=49$ and defined as
\begin{equation}
 \vect x^\top :=
 \left[\,
 h,\,
 \vect r_6^\top,\,
 \vect v^\top,\,
 \vect\omega^\top,\,
 \vect p_{\mathrm{feet}}^\top,\,
 \vect q^\top,\,
 \dot{\vect q}^{\,\top}
 \,\right].
 \label{eq:motion_state}
\end{equation}
Here, $h$ is the base height; $\vect r_6\in\mathbb R^6$ represents the orientation
of the base with yaw removed, to not include global orientation bias; $\vect v$ and $\vect\omega$ are base linear and angular
velocities; and $\vect p_{\mathrm{feet}}\in\mathbb R^{12}$ stacks the four
foot positions relative to the base. These orientation, velocity, and foot
quantities use yaw-normalized coordinates. The vectors $\vect q$ and
$\dot{\vect q}$ contain the 12 joint positions and velocities.

Across these sources, the corpus contains 53 trajectories, 40 Kine2Go policy rollouts, and 13 teleoperated recordings, comprising 59,247 frames (16.5 minutes of motion) sampled at 60~Hz, giving $\mat X_i\in\mathbb R^{120\times49}$ and a 2~s context. From these, we extract 10,661 overlapping two-second windows of $L=120$ frames, with successive window starts spaced five frames apart. The coarse annotations are
\mbox{$y_i\in\mathcal Y=\{\text{idle},\allowbreak\text{locomotion},\allowbreak
\text{jump},\allowbreak\text{sit},\allowbreak\text{lie},
\text{unknown}\}$}. All recordings participate in representation learning and clustering, with no decomposition train/test split.
Weak supervision from five broad behavior categories leads to a learned representation in which density clustering identifies 16 style groups.

Style discovery uses an ensemble of $M=10$ encoders with latent dimension $d=16$, loss weights \mbox{$(\lambda_x,\lambda_\Delta,\lambda_{\mathrm{inst}},\lambda_+,\lambda_-)=(1,0.5,1,0.05,0.2)$}, contrastive temperature $\tau=0.1$, and cosine-similarity threshold $\mu=0.2$; HDBSCAN uses excess-of-mass selection with minimum cluster size 80 and minimum samples 17.

The task command specifies target body-frame planar velocities and yaw rate:
\begin{equation}
 \vect c_t=
 [v_{x,t}^{\mathrm{cmd}},\,v_{y,t}^{\mathrm{cmd}},\,
  \omega_{z,t}^{\mathrm{cmd}}]^\top.
 \label{eq:command}
\end{equation}
The robot observation $\vect o_t\in\mathbb R^{46}$ is
\begin{equation}
 \vect o_t^\top :=
 \left[
 (\vect\omega_t^B)^\top\;
 (\vect g_t^B)^\top\;
 (\vect q_t-\vect q^{\mathrm{def}})^\top\;
 \dot{\vect q}_t^\top\;
 \vect a_{t-1}^\top\;
 \widehat h_t\;
 (\widehat{\vect v}_t^B)^\top
 \right].
 \label{eq:actor_observation}
\end{equation}
Here, $\vect\omega_t^B$ is base angular velocity, $\vect g_t^B$ is projected gravity, $\vect q^{\mathrm{def}}$ is the default joint configuration, $\dot{\vect q}_t$ is joint velocity, and $\vect a_{t-1}$ is the previous action. The estimates $\widehat h_t$ and $\widehat{\vect v}_t^B$ give the base
height and linear velocity. Superscript $B$ denotes body coordinates. 

The actor outputs $\vect a_t\in\mathbb R^{12}$ at 50~Hz while training physics runs at 200~Hz; a low-level PD controller tracks \mbox{$\vect q_t^{\mathrm{des}}=\vect q^{\mathrm{def}}+0.15\,\vect a_t$} with an action scale of $0.15$~rad and gains $K_p=30$, $K_d=0.5$. Height is measured using a downward-facing time-of-flight (ToF) sensor, while velocity is estimated using a contact-aided invariant extended Kalman filter (InEKF)~\cite{hartley2020inekf}. During policy training, the InEKF is driven by simulated IMU measurements, joint kinematics, and foot contacts.

The reference pools $\mathcal R_k$ of Eq.~\eqref{eq:reference_pool} are formed within each cluster--recording pair by ranking windows by cosine similarity to the cluster mean and greedily retaining anchors at least $L$ frames apart. This yields 335 non-overlapping references with stored consensus codes. During training, styles are sampled uniformly, their supporting recordings are balanced, and the selected reference $\vect r_t\in\mathcal R_k$ is held for 10~s or until reset.

The command term $r_t^{\mathrm{cmd}}$ of Eq.~\eqref{eq:composed_reward} is instantiated by a velocity-tracking score. We compare the body-frame velocities measured at $t+1$ with the command at time $t$,
\begin{equation}
 \begin{aligned}
  \vect e_t
  &={}
  \begin{bmatrix}
   v_{x,t+1}^{B} & v_{y,t+1}^{B} & \omega_{z,t+1}^{B}
  \end{bmatrix}^{\top}
  -\vect c_t,\\
  r_t^{\mathrm{cmd}}
  &={}
  \sum_{a\in\{x,y,\omega\}}
  \frac{\kappa_a}{1+(e_{a,t}/\beta_a)^2},
 \end{aligned}
 \label{eq:cauchy}
\end{equation}
where $e_{a,t}$ is the component of $\vect e_t$ on axis $a$, and $\kappa_a$ and $\beta_a$ are its weight and error scale. We use \mbox{$(\kappa_x,\kappa_y,\kappa_\omega)=(0.50,0.25,0.25)$}, $\beta_x=0.50$~m/s, \mbox{$\beta_y=0.20$}~m/s, and $\beta_\omega=0.50$~rad/s; the gate parameters are $(\nu,\delta_g,T_g)=(0.5,0.35,0.08)$, with post-ramp weights $w_{\mathrm{AMP}}=2$ and $w_{\mathrm{cmd}}=4$. The auxiliary term $r_t^{\mathrm{aux}}$ combines actuation and contact regularization with a reward matching the selected reference's base-height mean and variation.

\section{Experimental Evaluation}
\label{sec:controller_evaluation}

Our evaluation addresses three questions. First, whether weakly supervised consensus discovery yields a compact and kinematically coherent style repertoire, which we assess through ablations of the representation loss and comparisons against label-free and discrete alternatives (Sec.~\ref{subsec:coherence}). Second, whether a single style-conditioned policy attains command fidelity across the repertoire (Sec.~\ref{sec:commandability}) while preserving the motion characteristics of the selected reference as the command varies, including beyond the estimated recorded support of that style (Sec.~\ref{sec:style_expressibility}). Third, whether the resulting behavior persists outside the simulator used for training, which we examine through cross-simulator rollouts and onboard deployment (Sec.~\ref{sec:deployment}).

\subsection{Repertoire Coherence}\label{subsec:coherence}

Coherence requires distinguishable motion groups and consistent assignments. All comparisons use the same 53 recordings and 10,661 two-second windows, without a train/test split. We evaluate representation quality through repertoire size, total groups $K$ and effective count $K_{\mathrm{eff}}=\exp(-\sum_{k=0}^{K-1}p_k\log p_k)$, where $p_k=|\mathcal C_k|/\sum_j|\mathcal C_j|$; $K_{\mathrm{eff}}$ discounts sparsely occupied groups. We also consider coverage (fraction of non-rejected windows) and core switches to measure temporal fragmentation. Core switches count changes between successive assigned windows within two seconds, bridging short noise gaps but excluding transitions into or out of noise.

\emph{Balanced guidance prevents collapse.} Table~\ref{tab:decomposition-rewrite} evaluates variations of our representation loss, Eq.~\eqref{eq:representation_loss}. Without reconstruction, the repertoire collapses to just 3 groups ($K_{\mathrm{eff}}=1.20$), while uniform attraction reduces it to $7.40$. Both configurations sacrifice critical behavioral diversity. Conversely, the full CODES objective actively preserves a compact, semantically meaningful repertoire: 12 locomotion-dominant groups alongside distinct idle, jump, sit, and lie behaviors. By balancing attraction using confidence-weighted distributions, CODES avoids mode collapse while maintaining a highly reliable core (coverage of $72.8\%$).

\begin{table}[t]
 \centering
 \caption{Decomposition Ablations and Baselines}
 \label{tab:decomposition-rewrite}
 \begingroup
 \footnotesize
 \setlength{\tabcolsep}{1.5pt}
 \renewcommand{\arraystretch}{1.15}
 \begin{tabularx}{\columnwidth}{@{}>{\raggedright\arraybackslash}Xrrrrr@{}}
  \toprule
  Method & $K$ & $K_{\mathrm{eff}}$ & \shortstack{Loc.\\$K$} & \shortstack{Coverage\\(\%)} & \shortstack{Core switches\\(/min)} \\
  \midrule
  \rowcolor{gray!15}
  \textbf{CODES (ours)} & 16 & 10.19 & 12 & 72.8 & 1.36 \\
  \midrule
  \multicolumn{6}{@{}l}{\textit{Loss ablations}} \\
  No coarse ($\lambda_+=\lambda_-=0$) & 15 & 8.92 & 10 & 90.8 & 0.79 \\
  Separation only ($\lambda_+=0$) & 15 & 7.40 & 9 & 91.3 & 1.01 \\
  Uniform attraction ($\alpha_y=1$) & 7 & 3.42 & 7 & 95.9 & 0.00 \\
  No reconstruction ($\lambda_x=\lambda_\Delta=0$) & 3 & 1.20 & 3 & 93.6 & 0.00 \\
  \midrule
  \multicolumn{6}{@{}l}{\textit{Alternative representations}} \\
  Kinematic PCA & 39 & 30.00 & 30 & 84.1 & 3.02 \\
  TS2Vec consensus & 41 & 24.39 & 31 & 96.7 & 0.14 \\
  \bottomrule
 \end{tabularx}
 \endgroup
\end{table}

\emph{Consensus drives stability.} Our multi-encoder ensemble is fundamental to maintaining consistent temporal similarities. A single encoder yields unstable clustering ($4.6$ groups, $8.7\%$ coverage) and weak pairwise similarity correlation ($0.539$). Leveraging the full ensemble drives this correlation to $0.985$, demonstrating that consensus inherently stabilizes window-to-window similarities and produces robust partition boundaries.

\emph{Comparison with alternative representations.} We compare against Kinematic PCA, which clusters fixed posture and motion statistics; TS2Vec~\cite{yue2022ts2vec}, a label-free temporal representation; and VQ-VAE~\cite{han2024lifelike}, a discrete codebook alternative. The label-free Kinematic PCA and TS2Vec~\cite{yue2022ts2vec} baselines produce substantially larger repertoires, with 39--41 groups compared with 16 for CODES (Table~\ref{tab:decomposition-rewrite}). Kinematic assignments also switch more frequently ($3.02$ versus $1.36$ switches per minute). In contrast, the discrete VQ-VAE baseline~\cite{han2024lifelike} exhibits pronounced codebook underutilisation: only 17 of 256 codes are active, with $51.5\%$ of all windows assigned to a single code ($K_{\mathrm{eff}}=5.76$).

\emph{Superior kinematic coherence.} Table~\ref{tab:vq-coherence-rewrite} evaluates assignment coherence in a label-independent, 32-dimensional kinematic PCA space. Silhouette~\cite{rousseeuw1987silhouettes} measures within-group cohesion and between-group separation, while cross-recording AUC~\cite{fawcett2006roc} tests whether same-group motion characteristics recur across different recordings.

\begin{table}[t]
 \centering
 \caption{Coherence in a Common Kinematic Space}
 \label{tab:vq-coherence-rewrite}
 \begingroup
 \footnotesize
 \setlength{\tabcolsep}{3pt}
 \renewcommand{\arraystretch}{1.15}
 \begin{tabularx}{\columnwidth}{@{}Xrrr@{}}
  \toprule
  Method & \shortstack{Coverage\\(\%)} & \shortstack{Silhouette\\$\uparrow$} & \shortstack{Cross-recording\\AUC $\uparrow$} \\
  \midrule
  \rowcolor{gray!15}
  \textbf{CODES (ours)} & 72.8 & \bestresult{0.329} & \bestresult{0.832} \\
  No coarse terms & 90.8 & 0.264 & 0.826 \\
  VQ-VAE & 100.0 & -0.043 & 0.562 \\
  \midrule
  \multicolumn{4}{@{}l}{\textit{Coverage and balance controls}} \\
  \rowcolor{gray!15}
  CODES, full coverage & 100.0 & \bestresult{0.246} & \bestresult{0.790} \\
  VQ-VAE, balanced & 100.0 & 0.022 & 0.680 \\
  \bottomrule
 \end{tabularx}
 \endgroup
\end{table}

CODES achieves higher kinematic coherence than VQ-VAE, with a silhouette score of \(0.329\) and a cross-recording AUC of \(0.832\), compared with \(-0.043\) and \(0.562\), respectively.

To separate the effect of sample rejection from representation quality, we additionally assign every non-core window to its nearest CODES prototype, giving \(100\%\) coverage. Under this forced assignment, CODES retains a silhouette score of \(0.246\) and an AUC of \(0.790\), indicating that the improvement is not solely attributable to HDBSCAN rejection.

Ultimately, CODES provides a compact, kinematically differentiated repertoire. By combining coarse guidance with multi-encoder consensus, it yields robust representations that strictly outperform standard label-free and discrete alternatives, paving the way for reliable downstream control. Visualizations of CODES and baseline styles are on the project website.

\begin{figure*}[!t]
 \centering
 \includegraphics[width=\textwidth]{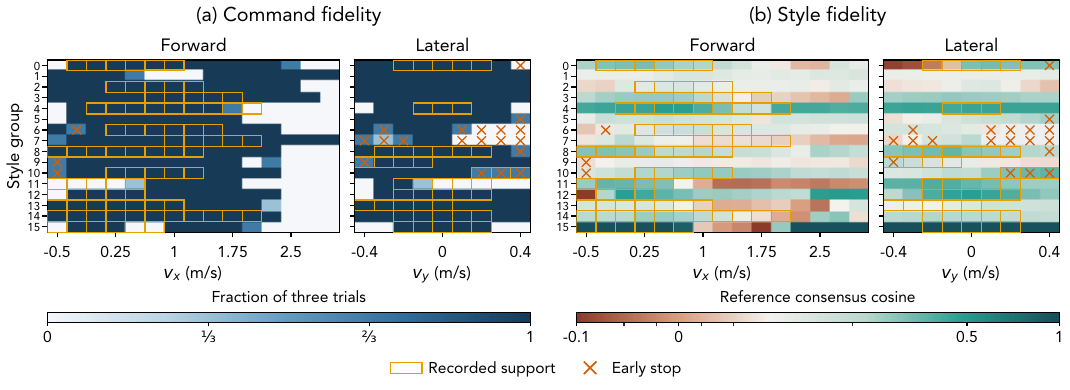}
 \caption{\textbf{Command and style fidelity across forward and lateral velocity sweeps, with the other commanded components set to zero.} (a) Fraction of three trials satisfying the completion and tracking criteria. (b) Mean reference consensus cosine, shown on an asinh scale clipped below $-0.1$. In reviewed rollouts, values near $0.5$ usually retain recognizable reference posture and limb coordination, whereas values near $0$ show weaker resemblance; this learned score is not a calibrated perceptual distance. Orange outlines show estimated recorded support; crosses mark early termination. Tracking extends beyond support, but reference resemblance decreases.}
 \label{fig:model-a-commandability}
 \label{fig:model-a-reference-cosine}
\end{figure*}

\begin{figure*}[!t]
 \centering
 \includegraphics[width=\textwidth]{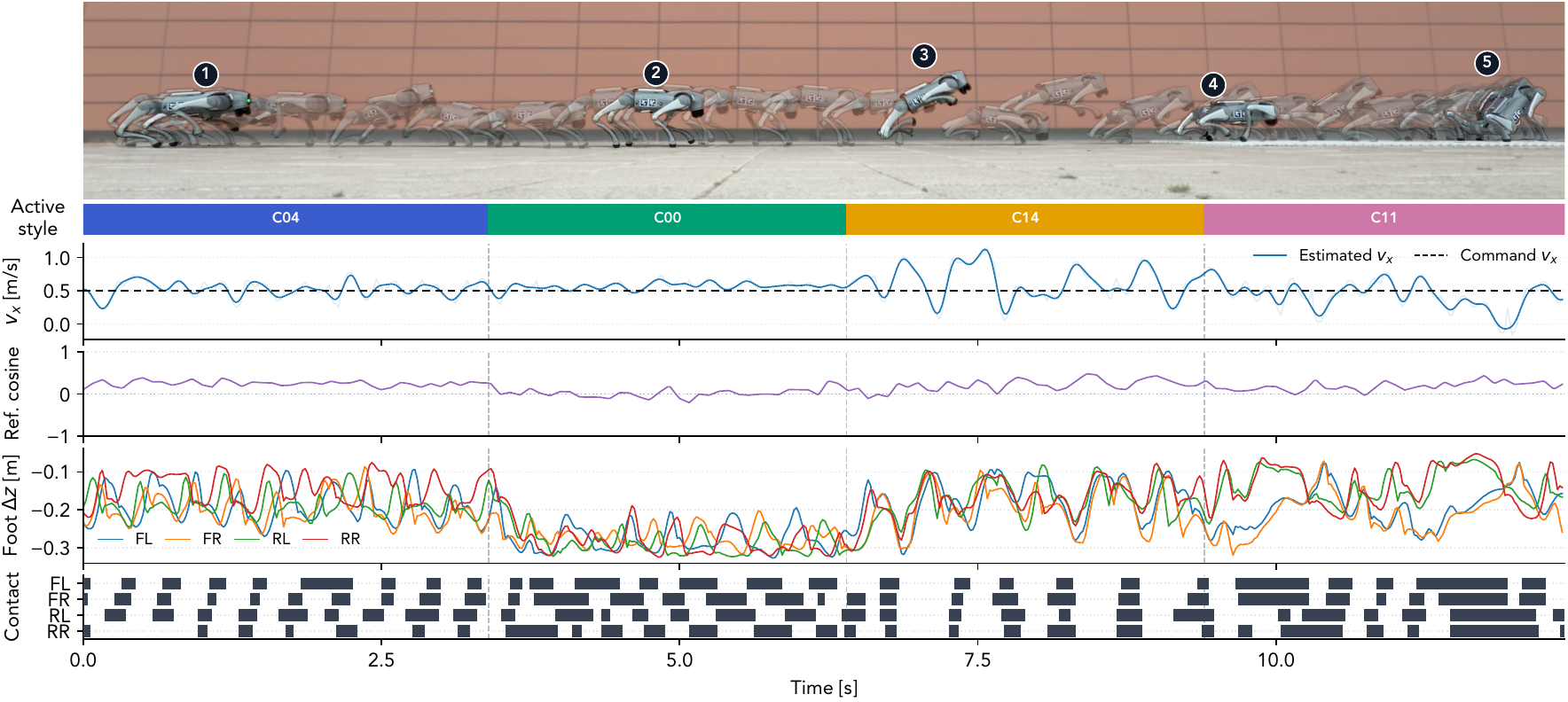}
  \caption{\textbf{Style switching on hardware under a fixed task command.} Deployment of the CODES policy on the Unitree Go2 at a constant $0.5$~m/s forward command with zero lateral and yaw-rate components, while the selected reference switches among C04, C00, C14, and C11. The colored band indicates the active style, and dashed lines the switch instants. The policy, reference bank, and state estimation run onboard, without external motion capture or offboard inference. The top row shows a time-lapse composite of the traversal, with numbered markers at successive instants of the run. The panels below report, from top to bottom, the estimated and commanded forward velocity $v_x$; the reference consensus cosine of Eq.~\eqref{eq:evaluation-reference-cosine} with respect to the active reference; the base-relative height $\Delta z$ of the four feet (FL, FR, RL, RR); and the kinematically inferred contact states. The command is identical across all four segments, while foot-swing amplitude, contact timing, and base posture differ, indicating that the reference changes how the command is executed rather than which command is tracked. Videos of the hardware execution are provided on the project website.}
 \vspace{-1mm}
 \label{fig:real-deployment-rewrite}
\end{figure*}

\subsection{Command Fidelity}
\label{sec:commandability}

Controller comparisons test coarse guidance through the no-coarse-terms ablation and discrete conditioning through VQ-VAE. All policies use the same training protocol: 20,000 iterations from scratch with 4,096 environments and the PPO, reward, observation, and reference-sampling setup of Section~\ref{sec:case_study}. Hidden layers are unchanged; input widths accommodate 160-dimensional consensus or 32-dimensional quantized codes. Each policy uses its own decomposition's reference bank. Baseline comparisons are conducted only in Isaac Sim; cross-simulator and hardware evaluation is limited to CODES.

\emph{Tracking measure and trials.} Command fidelity is measured by planar and yaw-rate RMSE over trial $i$'s scored interval $\mathbf T_i$:
\begin{equation}
 \begin{aligned}
 E_i^{xy}&=\sqrt{\frac{\sum_{t\in\mathbf T_i}(e_{x,i,t}^2+e_{y,i,t}^2)}{|\mathbf T_i|}},\\
 E_i^{\omega}&=\sqrt{\frac{\sum_{t\in\mathbf T_i}e_{\omega,i,t}^2}{|\mathbf T_i|}}.
 \end{aligned}
 \label{eq:evaluation-command-error}
\end{equation}
where $e_{x,i,t}$, $e_{y,i,t}$, and $e_{\omega,i,t}$ are measured minus commanded body-frame velocities, as in Eq.~\eqref{eq:cauchy}. The command distance in Eq.~\eqref{eq:commandability} is
\begin{equation}
 d_{\mathrm{cmd}}(\Bar{\vect u}_i,\vect u_i)
 =\max\!\left\{
 \frac{E_i^{xy}}{0.30\ \mathrm{m\,s^{-1}}},
 \frac{E_i^{\omega}}{0.40\ \mathrm{rad\,s^{-1}}}
 \right\}.
 \label{eq:evaluation-command-distance}
\end{equation}
With $\varepsilon_{\mathrm{cmd}}=1$, a \emph{successful trial} completes its full duration with planar RMSE at most $0.30$~m/s and yaw-rate RMSE at most $0.40$~rad/s. Continuous RMSE and completion are also reported separately.

Each dictionary entry is tested through one reference at 135 command pairs: forward velocity from $-0.5$ to $3.0$~m/s in $0.25$~m/s steps, lateral velocity from $-0.4$ to $0.4$~m/s in $0.1$~m/s steps, and zero yaw rate. Each pair uses three reset seeds and $15$~s trials; $\mathbf T_i$ excludes the first $5$~s for settling.

\begin{table}[t]
 \centering
 \caption{Command Fidelity in Simulation against Baselines}
 \label{tab:controller-rewrite}
 \begingroup
 \footnotesize
 \setlength{\tabcolsep}{3pt}
 \renewcommand{\arraystretch}{1.15}
 \begin{tabularx}{\columnwidth}{@{}Xrrrr@{}}
  \toprule
  Dictionary & $K$ & \shortstack{Median $E^{xy}$\\(m/s) $\downarrow$} & \shortstack{Complete\\(\%) $\uparrow$} & \shortstack{Success\\(\%) $\uparrow$} \\
  \midrule
  \rowcolor{gray!15}
  \textbf{CODES (ours)} & 16 & 0.114 & 98.1 & 74.2 \\
  No coarse terms & 15 & \bestresult{0.106} & \bestresult{99.3} & \bestresult{89.8} \\
  VQ-VAE & 17 & 0.142 & 99.2 & 66.8 \\
  \bottomrule
 \end{tabularx}
 \endgroup
\end{table}

Figure~\ref{fig:model-a-commandability} shows two axis-aligned slices of the full command grid. The forward sweep varies $v_x$ with $v_y=\omega_z=0$, whereas the lateral sweep varies $v_y$ with $v_x=\omega_z=0$. Each cell gives the success fraction over three trials; crosses mark early termination. Orange outlines enclose $95\%$ of the smoothed recorded-velocity probability, weighting recordings equally. They indicate typical recorded commands, not success thresholds or policy-training boundaries.

CODES attains median planar RMSE $0.114$~m/s and $74.2\%$ success across 6,480 trials, versus $89.8\%$ success without coarse terms and $66.8\%$ for VQ-VAE (Table~\ref{tab:controller-rewrite}). In the forward sweep, tracking is strongest at low and moderate speeds, degrading for several references above approximately $2.25$~m/s. The no-coarse ablation therefore has stronger aggregate command metrics, including median planar RMSE of $0.106$~m/s. Each policy uses its own reference bank, so these results do not isolate a command-tracking tradeoff for identical requested motions.

\subsection{Style Fidelity Under Command Changes}
\label{sec:style_expressibility}

\emph{Reference consensus cosine} assesses style fidelity using the frozen encoders and the consensus similarity $G$ of Eq.~\eqref{eq:consensus}:
\begin{equation}
 G(\Bar{\mat X}_{i,t},\mat X_{r_i})
 =\widetilde{\vect z}(\Bar{\mat X}_{i,t})^\top\widetilde{\vect z}_{r_i}.
 \label{eq:evaluation-reference-cosine}
\end{equation}
Here $\Bar{\mat X}_{i,t}$ is the two-second rollout window ending at $t$, normalized as in discovery, and $\mat X_{r_i}$ is the target-robot window of reference $\vect r_i$. Higher cosine indicates closer resemblance; $1-G$ is a reference-distance proxy for Eq.~\eqref{eq:expressibility}. Scores are averaged over $\mathbf T_i$, excluding windows spanning a reference switch. Figure~\ref{fig:model-a-reference-cosine}(b) averages the three trial scores per cell, reporting similarity continuously rather than thresholding style success.

Within and beyond the estimated recorded support, respectively, mean cosine is $0.271$ and $0.156$ for the forward sweep, and $0.351$ and $0.219$ for the lateral sweep. The corresponding command-success rates are $87.8\%$ and $65.6\%$ forward, and $89.1\%$ and $78.9\%$ lateral. Thus, command tracking extends beyond recorded support in both directions, although reference resemblance decreases on average.

Consensus cosine reflects velocity as well as posture and coordination, so it is not command-independent. The physical traces in Figure~\ref{fig:real-deployment-rewrite} complement this measure by showing distinct postures and foot motions at a fixed forward command: the reference changes how the command is executed.

\subsection{Cross-Simulator and Hardware Execution}
\label{sec:deployment}

The policy, reference bank, and state estimation run onboard the robot's Jetson, without external motion capture or offboard inference. The policy is further fine-tuned for 2,000 iterations with joint targets clipped to $0.025$~rad within the physical limits, matching the safeguard used during deployment.

Figure~\ref{fig:real-deployment-rewrite} shows a completed Go2 mission at a fixed forward command of $0.5$~m/s. Switching through C04, C00, C14, and C11 changes posture and foot-motion amplitude and timing; C04 remains lower than C00 and C14. Onboard velocity estimation and kinematic contact inference are described in Section~\ref{sec:case_study}. A MuJoCo rollout with the same checkpoint and reference sequence exhibits corresponding motion changes, supporting cross-simulator transfer for these references.

\section{Conclusion}
\label{sec:limitations}

We presented CODES, a framework that discovers motion styles from heterogeneous recordings and conditions a robot controller on a selected style and an independently specified feasible command. Retargeting grounds discovery in executable motion, while consensus encoding and density clustering produce a repertoire without prescribing its size. On the Go2, CODES achieved higher common-space kinematic coherence than its no-coarse ablation and the VQ-VAE baseline. The controller tracked velocity commands beyond the styles' estimated recorded support and produced reference-dependent motion changes in simulation and hardware.

The evaluation also exposes limitations: the no-coarse controller achieved higher command success on its own reference bank, and reference consensus cosine declined outside recorded support. These findings motivate command-independent measures of style fidelity and automatic reference selection~\cite{fu2026diverse,han2024lifelike}. Future work should also examine transfer across robots and multi-robot coordination, where multiple group motion patterns may accomplish the same task but be difficult to specify analytically.

\bibliographystyle{IEEEtran}
\bibliography{reference}

\end{document}